\pdfoutput=1

\documentclass[11pt]{article}

\usepackage{acl}

\usepackage{times}
\usepackage{latexsym}

\usepackage[T1]{fontenc}
\usepackage[utf8]{inputenc}

\usepackage{microtype}

\usepackage{graphicx}

\usepackage{booktabs}
\usepackage{multirow}

\newcommand{\ascentpp}{\textsc{Ascent++}}
\newcommand{\conceptnet}{\textsc{ConceptNet}}
\newcommand{\comet}{\textsc{Comet}}
\newcommand{\atomic}{\textsc{Atomic}}

\newcommand{\triple}[1]{\emph{$\langle$#1$\rangle$}}

\renewcommand{\paragraph}[1]{\smallskip\noindent\textbf{#1.\mbox{\ \ }}}

\title{Materialized Knowledge Bases from Commonsense Transformers}

\author{Tuan-Phong Nguyen \\
  Max Planck Institute for Informatics \\
  Saarland Informatics Campus \\
  Saarbrücken, Germany \\ 
  \texttt{tuanphong@mpi-inf.mpg.de} \And
  Simon Razniewski \\
  Max Planck Institute for Informatics \\
  Saarland Informatics Campus \\
  Saarbrücken, Germany \\
  \texttt{srazniew@mpi-inf.mpg.de}}

\begin{document}
\maketitle
\begin{abstract}
Starting from the \comet{} methodology by \citet{bosselut2019comet},
generating commonsense knowledge 
% directly from pre-trained language models 
from commonsense transformers
has recently received significant attention.
Surprisingly, up to now no materialized resource of commonsense knowledge generated this way is publicly available. This paper fills this gap, and uses the materialized resources to perform a detailed analysis of the potential of this approach in terms of precision and recall. Furthermore, we identify common problem cases, and outline use cases enabled by materialized resources. 
We posit that the availability of these resources is important for the advancement of the field, as it enables an off-the-shelf-use of the resulting knowledge, as well as further analyses on its strengths and weaknesses.
\end{abstract}

\section{Introduction}

Compiling comprehensive collections of commonsense knowledge (CSK) is an old dream of AI. Besides attempts at manual compilation~\cite{liu2004conceptnet,lenat1995cyc,atomic} and text extraction~\cite{schubert2002can,webchild,mishra2017domain,quasimodo,ascentpp}, commonsense knowledge compilation from pretrained language models~\cite{bosselut2019comet,comet-atomic-2020,west2021symbolic} has recently emerged.
% Pre-trained language models have shaken up NLP in general, and also shown promising performance in generating commonsense assertions, based on fine-tuning on existing corpora of commonsense assertions.
In \citeyear{bosselut2019comet}, \citeauthor{bosselut2019comet} introduced \textit{Commonsense Transformers} (\comet{}), an approach for fine-tuning language models on existing corpora of commonsense assertions. 
These models have shown promising performance in generating commonsense assertions after being trained on established human-authored commonsense resources such as \atomic~\cite{atomic} and \atomic$^{20}_{20}$~\cite{comet-atomic-2020}.

More recently, \citet{west2021symbolic} extracts commonsense assertions from a general language model, GPT-3~\cite{GPT3}, using simple prompting techniques. Surprisingly, using this machine-authored commonsense corpus to fine-tune \comet{} helps it outperform GPT-3, which is 100x larger in size, in terms of commonsense capabilities.

Despite the prominence of this approach (the seminal \comet{} paper~\cite{bosselut2019comet} receiving over 300 citations in just two years), to date, no resource containing commonsense knowledge compiled from any \comet{} model is publicly available. As compilation of such a resource is a non-trivial endeavour, this is a major impediment to research that aims to understand the potentials of the approach, or intends to employ its outputs in downstream tasks.

This resource paper fills this gap. We fine-tune the \comet{} pipeline on two established resources of concept-centric CSK assertions, \conceptnet{} \cite{speer2017conceptnet} and \ascentpp{} \cite{ascentpp}, and execute the pipeline for 10K prominent subjects. 
Unlike the \atomic{} resources, which were used to train \comet{} in \cite{bosselut2019comet,comet-atomic-2020} and have their main focus on events and social interactions, the two resources of choice are mostly about general concepts (e.g., \textit{lions can roar}, or \textit{a car has four wheels}).
Furthermore, as those two resources were constructed using two fundamentally different methods, crowdsourcing and web text extraction, it enables us to discover potentially different impacts they have on the \comet{} models.

By taking the top-10 inferences for each subject-predicate pair, we obtain four resources, \conceptnet{} (GPT2-XL, BART) and \ascentpp{} (GPT2-XL, BART), containing 900K to 1.4M ranked assertions of CSK. We perform a detailed evaluation of the intrinsic quality, including fine-grained precision (typicality and saliency) and recall of each resource, derive qualitative insights into the strengths and weaknesses of the approach, and highlight extrinsic use cases enabled by the resources.

\pagebreak

Our contributions are:
\begin{enumerate}
    \item The materialization of the \comet{} approach for two language models (GPT2-XL, BART) on two concept-centered commonsense knowledge bases (\conceptnet{}, \ascentpp{});
    \item Quantitative and qualitative evaluations of the resulting resources in terms of precision, recall and error categories, showing that in terms of recall, \comet{} models outperform crowdsourced construction and are competitive with web text extraction, while exhibiting moderate gaps in terms of precision to both;
    \item Illustrative use cases of the materialized resources in statement aggregation, join queries, and search.
\end{enumerate}
The materialized resources, as well as an interactive browsing interface, are available at\linebreak {\small \url{https://ascentpp.mpi-inf.mpg.de/comet}}.
% The materialized resources can be downloaded at {\small \url{https://www.dropbox.com/s/wibimhbgire3jwc/comet.tar.gz?dl=0}}.
% A web interface for browsing the resources will be made publicly available (see a screenshot of the interface in Figure~\ref{fig:interface}).

\section{Related work}

Early approaches at CSK compilation relied on expert knowledge engineers \cite{lenat1995cyc} or crowdsourcing \cite{liu2004conceptnet}, and the latter approach has recently been revived \cite{atomic}. To overcome scalability limitations of manual compilation, text extraction is a second popular paradigm. Following early attempts on linguistic corpora \cite{mishra2017domain}, increasingly approaches have targeted larger text corpora like Wikipedia, book scans, or web documents \cite{webchild,quasimodo,ascentpp,ascent}, to build CSK resources of wide coverage and quality.

Recently, both approaches have been complemented by knowledge extraction from pre-trained language models: 
Language models like BERT~\cite{devlin2019bert} or GPT~\cite{radford2019language, GPT3} have seen millions of documents, and latently store associations among terms.
While \citet{west2021symbolic} used prompting to extract symbolic CSK from GPT-3,
\citet{bosselut2019comet} proposed to tap this knowledge by supervised learning: 
The language models are fine-tuned on statements from existing knowledge resources, e.g., trained to predict the object \textit{Africa} when given the subject-predicate pair \textit{elephant, AtLocation}, based on the ConceptNet triple \triple{elephant, AtLocation, Africa}. 
After training, they can be used to predict objects for unseen subject-predicate pairs, e.g., locations of wombats.

The approach gained significant attention, and variants are employed in a range of downstream tasks, e.g., commonsense question answering \cite{bosselut2019dynamic}, commonsense explanation~\cite{semeval-csk-explanation}, story generation \cite{guan2020knowledge}, or video captioning~\cite{fang2020video2commonsense}.

Yet, to date, no materialized knowledge resource produced by any \comet{} model is available (\textsc{AutoTOMIC} from \cite{west2021symbolic} being based on prompting GPT-3). The closest to this is a web interface hosted by the AllenAI institute at {\small \url{https://mosaickg.apps.allenai.org/model_comet2020_entities}}. However, this visualizes only predictions for a single subject, making, e.g., aggregations or count impossible, and only shows top-5 predictions, and without scores.

\section{Methodology}

We follow the implementations in the official code repository\footnote{\url{https://github.com/allenai/comet-atomic-2020/}} of the \textsc{Comet-Atomic}$_{20}^{20}$ project~\cite{comet-atomic-2020}
% , filling however some gaps in terms of (what was it - about special tokens?), and 
to compute assertions, and decide on output thresholds.

\paragraph{Training CSKBs}
We use two established concept-centered commonsense knowledge bases (CSKBs), \conceptnet{} 5.7~\cite{speer2017conceptnet} and \ascentpp{}~\cite{ascentpp} as training resources, considering 13 CSK predicates from each of them: \textit{AtLocation}, \textit{CapableOf}, \textit{Causes}, \textit{Desires}, \textit{HasA}, \textit{HasPrerequisite}, \textit{HasProperty}, \textit{HasSubevent}, \textit{MadeOf}, \textit{MotivatedByGoal}, \textit{PartOf}, \textit{UsedFor} and \textit{ReceivesAction}.

\begin{enumerate}
    \item \conceptnet{}~\cite{speer2017conceptnet} is arguably the most widely used CSKB, built by crowdsourcing. \conceptnet{} 5.7 is its lastest version\footnote{\url{https://github.com/commonsense/conceptnet5/wiki/Downloads}}, consisting of 21 million multilingual assertions, spanning CSK as well as general linguistic and taxonomic knowledge. We retain English assertions only, resulting in 207,210 training assertions for the above-mentioned predicates. 
    \item \ascentpp{}~\cite{ascentpp} is a project aiming for automated CSK extraction from large-scaled web contents based on open information extraction (OpenIE) and judicious cleaning and ranking approaches. The \ascentpp{} KB consists of 2 million English CSK assertions for the 13 mentioned predicates.
\end{enumerate}

% \begin{enumerate}
%     \item ConceptNet 5.7 (only CSK part): 207K assertions from \cite{speer2017conceptnet}, considering the 13 CSK relations.
%     \item \ascentpp: 2M assertions from the open schema CSK resource \cite{ascentpp}, mapped to the ConceptNet schema by a multi-class classifier followed by a rule-based model for object modification.
% \end{enumerate}

% We query the fine-tuned models for the 10k subjects in ConceptNet with the most assertions, for 13 relations: AtLocation, CapableOf, Causes, Desires, HasA, HasPrerequisite, HasProperty, HasSubevent, MadeOf, MotivatedByGoal, PartOf, UsedFor and ReceivesAction.

% For each subject-relation pair, we use beam search to obtain completions. We retain all completions with (perplexity?) greater ..., this threshold being decided based on a small sample of withheld data.

% \sr{CAN WE HAVE TWO SETTINGS? ONCE SAME SIZE AS INPUT, ONCE BIGGER? THEN HAVE A CONCEPTNET-COMET-SAMESIZE, and a CONCEPTNET-COMET-EXTENDED etc.?}

\paragraph{Language models}
We consider two autoregressive language models (LMs) that were also used in the original \comet{} paper, GPT2-XL~\cite{radford2019language} and BART~\cite{lewis2019bart}.

\paragraph{Materialization process}
We query the fine-tuned \comet{} models for 10,926 subjects in \conceptnet{} which have at least two assertions for the 13 CSK predicates. 
For each subject-predicate pair, we use beam search to obtain completions, with different configurations (see Table~\ref{tab:configs}) for BART and GPT2-XL, following the parameters specified in the published code repository and models. 
We retain the top-10 completions for each subject-predicate pair, with their \textit{beam scores} (i.e., sum of log softmax of all generated tokens) returned by the \textit{generate} function\footnote{\url{https://huggingface.co/docs/transformers/main/en/main\_classes/text\_generation\#transformers.generation\_utils.GenerationMixin.generate}} of the Transformers library~\cite{transformers}.

\begin{figure*}[t]
    \centering
    \frame{\includegraphics[width=\textwidth]{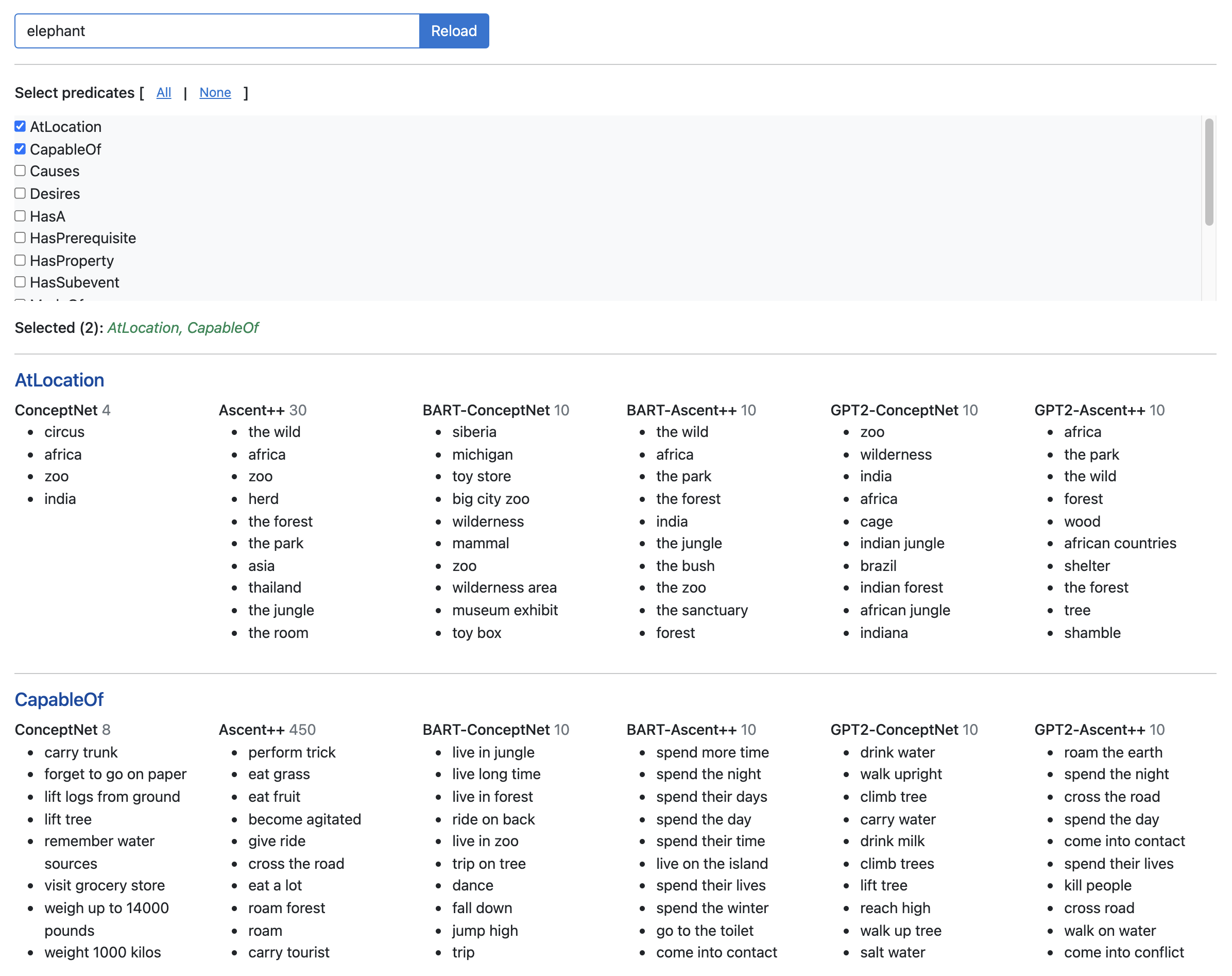}}
    \caption{Web interface showing top-10 assertions per predicate in six CSK resources. The number in grey next to a CSKB indicates the total number of assertions for the corresponding subject-predicate pair in the KB.}
    \label{fig:interface}
\end{figure*}

\paragraph{Output}
The resulting resources, \conceptnet{} (GPT2-XL, BART) and \ascentpp{} (GPT2-XL, BART), contain a total of 976,296 and 1,420,380 and 1,271,295 and 1,420,380 assertions after deduplication, respectively, as well as their corresponding beam scores. 
All are available for browsing, as well as for download, at {\small \url{https://ascentpp.mpi-inf.mpg.de/comet}} (see screenshot of browsing interface in Figure~\ref{fig:interface}). 
% \sr{this link is problematic for double blind review (if the workshop is so). An alternative could be to provide the resources in dropbox, and show only a short video of the browsing interface?}

\begin{table}[t]
    \centering
    \small
    \begin{tabular}{lrr}
        \toprule
        \textbf{Parameter} & \textbf{GPT2-XL} & \textbf{BART} \\
        \midrule
        num\_beams & 10 & 10 \\
        temperature & 1.0 & 1.0 \\
        top\_p & 0.9 & 1.0 \\
        repetition\_penalty & 1.0 & 1.0 \\
        max\_length & 16 & 24 \\
        no\_repeat\_ngram\_size & 0 & 3 \\
        early\_stopping & True & True \\
        do\_sample & False & False \\
        \bottomrule
    \end{tabular}
    \caption{Configurations for beam-search decoders.}
    \label{tab:configs}
\end{table}

\section{Analysis}

We perform three kind of analyses: (1) a quantitative evaluation of the intrinsic quality of the assertions, based on crowdsourcing, (2) a qualitative evaluation that outlines major strengths and weaknesses, and (3) an illustration of use cases enabled by both resources.

\subsection{Quantitative evaluation}

% ilievski only evaluates on downstream tasks in CSKG, dimensions of CSK. in consolidating CSK even less.
%comet-atomic-2020 mostly on transfer - split KG into train, test, see whether can predict other parts from it - in other words, tests redundancy (although no leakage test..)
%comet mostly on bleu-2, and perplexity, and novelty of SPO and O - automated metrics 
% - again, bleu is based on train/test-split - tests capability of model relative to source data, not quality of output...

%original paper only evaluated plausibility of top-1 triple per s-p-pair (greedily decoded setting) - 92\% - but plausibility a weak concept.

The original paper \cite{bosselut2019comet} only evaluated the top-1 triple per subject-predicate pair. Furthermore, it solely evaluated triples by plausibility, which is a necessary, but only partly a sufficient criterion for being considered commonsense \cite{chalier2020joint}.

In the following, we evaluate samples from the generated resources along two \textit{precision} dimensions, typicality (top-100 assertions per subject) and saliency (top-10 assertions per subject). We also evaluate \textit{recall}, by measuring the degree to which each resource covers the statements in a human-generated ground truth.

\paragraph{Precision: Typicality and saliency}
Following~\citet{quasimodo,ascentpp}, we assess assertions in the CSK resources along two precision dimensions: \textit{typicality} and \textit{saliency}, which measure the degree of truth and the degree of relevance of assertions, respectively. We use the Amazon Mechanical Turk (AMT) platform to obtain human judgements. Each dimension is evaluated based on a 4-point Likert scale and an option for \textit{no judgement} if the annotator is not familiar with the concepts. Assertions are transformed into human-readable sentences using the templates introduced by \citet{comet-atomic-2020}. Each assignment is done by three different workers. Following~\citet{comet-atomic-2020}, any CSK assertion that receives the two higher scores in the Likert scale is labelled as \textit{Typical} or \textit{Salient}, and the two lower scores as \textit{Untypical} or \textit{Unsalient}. The final judgements is based on majority vote.

In terms of sampling process, for typicality, we draw 500 assertions from each resource when restricting to top-100 assertions per subject. For saliency, we pick 500 random samples from the pool of top-10 assertions per subject.

Results are reported in the left part of Table~\ref{tab:csk-eval}. We see a significant drop in the quality of assertions in the LM-based generations compared to the training resources. In terms of the neural models, for both training CSKBs, the BART models demonstrate better typicality than the GPT2-XL ones. Assertions in BART-\ascentpp{} also have significantly better saliency than in GPT2-XL-\ascentpp{}. Interestingly, BART-\conceptnet{} is nearly on par with \ascentpp{} on both metrics.

\begin{table*}[t]
\centering
\small
\begin{tabular}{rrrrrrrrr}
    \toprule
    \multirow{2}{*}{\textbf{Resource}} & \multicolumn{2}{c}{\textbf{Typicality@100}} & \multicolumn{2}{c}{\textbf{Saliency@10}} & \multicolumn{3}{c}{\textbf{Recall@100}} & \textbf{Size@100} \\
    \cmidrule(l){2-3} \cmidrule(l){4-5} \cmidrule(l){6-8} \cmidrule(l){9-9}
     & \textbf{Typical} & \textbf{Untypical} & \textbf{Salient} & \textbf{Unsalient} & \textbf{t=0.96} & \textbf{t=0.98} & \textbf{t=1.00} & \textbf{\#triples} \\
    \cmidrule{1-1} \cmidrule(l){2-3} \cmidrule(l){4-5} \cmidrule(l){6-8} \cmidrule(l){9-9}
    \ascentpp{} & \textbf{78.4} & \textbf{11.0} & \textbf{62.8} & \textbf{34.6} & \textbf{8.9} & \textbf{7.9} & \textbf{4.6} & 202,026  \\
    GPT2-XL-\ascentpp{} & 57.2 & 27.4 & 37.2 & 58.4 & 6.0 & 4.9 & 2.6 & 1,091,662  \\
    BART-\ascentpp{} & 69.8 & 17.4 & 50.6 & 42.6 & 2.6 & 1.9 & 1.0 & 1,092,600  \\
    \cmidrule{1-1} \cmidrule(l){2-3} \cmidrule(l){4-5} \cmidrule(l){6-8} \cmidrule(l){9-9}
    \conceptnet{} & \textbf{93.6} & \textbf{3.6} & \textbf{80.0} & \textbf{16.8} & 2.3 & 1.7 & 0.9 & 164,291  \\
    GPT2-XL-\conceptnet{} & 66.6 & 21.4 & 63.8 & 32.6 & \textbf{9.0} & \textbf{7.3} & \textbf{3.8} & 967,343  \\
    BART-\conceptnet{} & 72.6 & 17.0 & 63.4 & 33.4 & 5.3 & 3.7 & 1.0 & 1,092,600  \\
    \bottomrule
\end{tabular}
\caption{Intrinsic evaluation (Typicality, Saliency and Recall - \%) and size of CSK resources.}
\label{tab:csk-eval}
\end{table*}

\paragraph{Recall}
We reuse the CSLB dataset~\cite{devereux2014centre} that was processed by~\citet{ascentpp} as ground truth for recall evaluation. The CSLB dataset consists of 22.6K human-written sentences about property norms of 638 concepts. To account for minor reformulations, following \citet{ascentpp}, we also use embedding-based similarity to match ground-truth sentences with statements in the CSK resources. 
We specifically rely on precomputed SentenceTransformers embeddings~\cite{sbert}.
We also restrict all CSK resources to top-100 assertions per subject. 

\begin{figure}[t]
    \centering
    \includegraphics[width=\columnwidth, trim =1cm 0 1.5cm 1.2cm,clip]{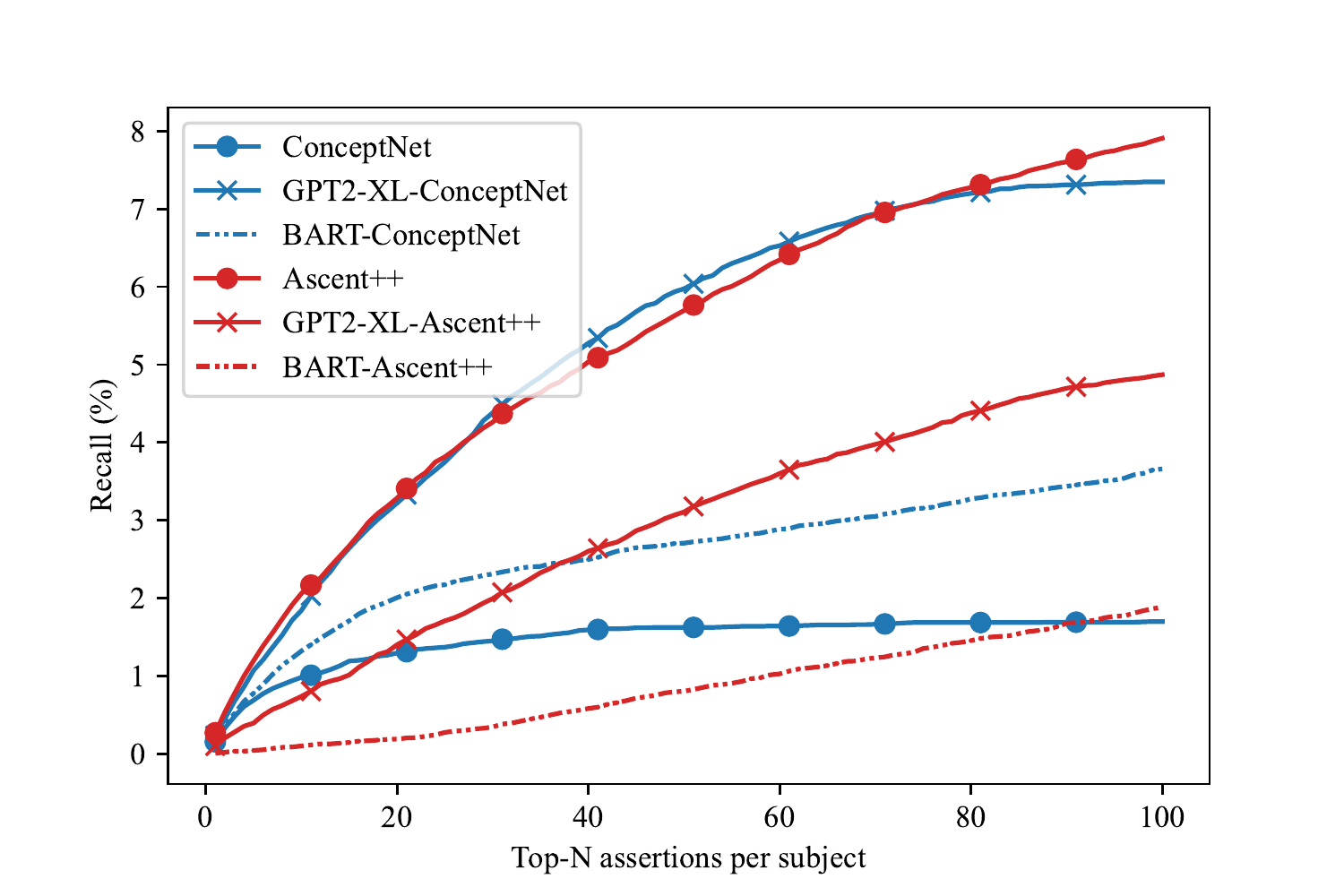}
    \caption{Resource recall in relation to resource size, at similarity threshold $t=0.98$. 
    % \sr{Modify color coding so that COMET-derived resources show coherence to base resources (e.g., dashed in same color, or related shades of base resource etc.}
    }
    \label{fig:recal-vs-size}
\end{figure}

The evaluation results are shown in the right part of Table~\ref{tab:csk-eval}, where we report recall at similarity thresholds $0.96$, $0.98$ and $1.0$, as well as resource size. We also plot the recall values at different top-N assertions per subject in Figure~\ref{fig:recal-vs-size} with similarity threshold $t=0.98$.
% \\
% - ascent does not benefit but conceptnet does \\
% - gpt2 is now the winner
%
As one can see, \ascentpp{} outperforms both \comet{} models trained on it even though it is significantly smaller. We see opposite results with the \conceptnet{}-based resources, where the \comet{} models generate resources of better coverage than its training data. Our presumption is that the LMs profits more from manually curated resources like \conceptnet{}, but hardly add values to resources that were extracted from the web, as LMs have not seen fundamentally different text.
Furthermore, in contrast to precision, GPT2-XL models have better results than BART models in terms of recall, on both input CSKBs.

% \begin{table}[t]
%     \centering
%     \small
%     \begin{tabular}{lrrrr}
%     \toprule
%     \textbf{Resource} & \textbf{t=0.96} & \textbf{t=0.98} & \textbf{t=1.00} & \textbf{\#triples}\\
%     \midrule
%     \ascentpp{} & \textbf{8.9} & \textbf{7.9} & \textbf{4.6} & 202,026 \\
%     \ascentpp{} (GPT2-XL) & 6.0 & 4.9 & 2.6 & 1,091,662 \\
%     \ascentpp{} (BART) & 2.6 & 1.9 & 1.0 & 1,092,600 \\
%     \midrule
%     \conceptnet{} & 2.3 & 1.7 & 0.9 & 164,291 \\
%     \conceptnet{} (GPT2-XL) & \textbf{9.0} & \textbf{7.3} & \textbf{3.8} & 967,343 \\
%     \conceptnet{} (BART) & 5.3 & 3.7 & 1.0 & 1,092,600 \\
%     \bottomrule
%     \end{tabular}
%     \caption{Recall@100 (\%) and size of CSK resources.}
%     \label{tab:recall-evaluation}
% \end{table}

\subsection{Qualitative observations}

% Based on sampling 50 statements labelled as not salient (by crowd workers), 50 labelled not as typical (from ConceptNet-GPT). Then assigning to error categories
%  - subject/object mix
%  - vague co-occurrence
%  - syntactically wrong (e.g., nouns instead of activities)?
%  - named entities
%  - overly specific
% \ph{Not interesting results. A lot of non-factual generations, but most of them do not fall into these categories. See the samples labelled as negative at \url{https://docs.google.com/spreadsheets/d/19KFzlQo_F8i-6ZzyclEqYwDz7flSilnH7Slqx5b6DR8/edit?usp=sharing}.}

% Other observations:
%  - Redundancy! (possibly with numbers: from top-10 statements of a few subjects, how many are redundant)
%  - Difference GPT vs. BART: BART is more redundant than the other, so lower R, but higher P (note: this might change with different parameter settings, but several were tried, and these ones, which are taken from the literature, performed best).
 
LMs have the strength to generate an open-ended set of objects, even for subjects seen rarely or not at all in the training data. 
For example, while \conceptnet{} stores only one location for \textit{rabbit}: \textit{``a meadow''}, both BART- and GPT2-XL-\conceptnet{} can generalize to other correct locations, such as \textit{wilderness}, \textit{zoo}, \textit{cage}, \textit{pet store}, etc.
In the recall evaluation, we pointed out that \conceptnet{}, a manually-built CSK resource with relatively small size, considerably benefits from LMs generations as they improve the coverage of the resource substantially. 

However, as indicated in the precision evaluation, LM generations are generally of lower precision than those in the training data. Common error categories we observe are:
\begin{itemize}
    \item \textbf{Co-occurrence misreadings:} LMs frequently predict values that merely frequently co-occur, e.g., \triple{locomotive, atLocation, bus stop}, \triple{running, capableOf, put on shoes}, \triple{war, desires, kill people}, \triple{supermarket, capableOf, buy milk}.
    \item \textbf{Subject-object-copying}: LMs too often repeat the given subject in predictions. For instance, 45 of 130 objects generated by BART-\conceptnet{} for the subject \textit{chicken} also contain \textit{chicken}, such as \triple{chicken, CapableOf, kill/eat/cook chicken} or \triple{chicken, UsedFor, feed chicken}.
    % \item \textbf{Vague co-occurrence}:
    % \item \textbf{Overly specific objects}:
    \item \textbf{Quantity confusion}: LMs struggle to distinguish quantities. For example, GPT2-XL-\conceptnet{} generates that \textit{bike} has \textit{four wheels} (top-1 prediction), and then also \textit{two wheels} (rank 3), \textit{three wheels} (rank 4) and \textit{twelve wheels} (rank 5). The weakness of dealing with numbers is known as a common issue of embeddings-based approaches \cite{numbers-embeddings}.
    \item \textbf{Redundancy}: Generated objects often overlap, bloating the output with redundancies. Most common are repetitions of singular/plural nouns, e.g., the top-2 generations by BART-\conceptnet{} for \textit{doctor-CapableOf}: \textit{``visit patient''} and \textit{``visit patients''}. Redundancies also include paraphrases, e.g., \triple{doctor, CapableOf, visit patients / see patients}; or \triple{doctor, CapableOf, prescribe medication / prescribe drug / prescribe medicine} (GPT2-XL-\ascentpp{} generations). Clustering might alleviate this issue \cite{ascentpp}. % Furthermore, we observe that BART generally makes more redundancies than GPT2-XL, which could be the reason why BART has lower recall but higher precision.
    % This might change with different parameter settings for the beam search decoder, but after several tries, and these ones (Table~\ref{tab:configs}), which are taken from the literature, performed best.
\end{itemize}

\subsection{Downstream use of materialized resources}

Beyond systematic evaluation, materialized resources enable a wide set of downstream use cases, for example context-enriched zero-shot question answering~\cite{petroni2020context}, or KB-based commonsense explanation~\cite{semeval-csk-explanation}.
We exemplarily illustrate four enabled types of basic analyses, (1) frequency aggregation, (2) join queries, (3) ranking and (4) text search.

% Perform some count query - how many objects eat ABC vs. how many eat DEF.

% Observe most frequent locations of items in general.

% Look for cycles - animals that eat each other.

% Other use case: Use as context in CSQA, following petroni approach - just mention this option. Use as context in SEMEVAL task. Etc.

\paragraph{Frequency aggregation}
Materialized resources enable to count frequencies. In Table~\ref{tab:common-objects}, we demonstrate the three most common objects for each predicate in the GPT2-XL-\conceptnet{} resource. Interestingly, the third most common location of items in the KB is \textit{``sock drawer''}, which is only ranked as the 190\textsuperscript{th} most common location in \conceptnet{}. Similarly, the top-3 objects for \textit{CapableOf} in the generated KB rarely occur the training data.

\paragraph{Join queries}
One level further, materialized knowledge enables the construction of join queries. For example, 
% concerning diets, 
we can formulate conjunctive queries like:

\begin{itemize}
    % \item The most common diets \triple{$<$subject$>$, CapableOf, eat ...} are \textit{worms} (29), \textit{mice} (17), \textit{soup} (17), \textit{meat} (16) and \textit{grass} (15).
    \item Animals that eat themselves include \textit{chicken}, \textit{flies}, \textit{grasshopper}, \textit{mice}, \textit{penguin}, \textit{worm}.
    \item The most frequent subevents of subevents are: \textit{breathe}, \textit{swallow}, \textit{hold breath}, \textit{think}, \textit{smile}.
    \item The most common parts of locations are: \textit{beaches}, \textit{seeds}, \textit{lot of trees}, \textit{peel}, \textit{more than one meaning}.
\end{itemize}

\paragraph{Ranking}
Since statements in our materialized resources come with scores, it becomes possible to locally and globally rank assertions, or to compare statements pairwise. For example, in GPT2-XL-\conceptnet{}, the triple \triple{librarian, AtLocation, library}, which is at rank 140, has a score of $-0.048$, which is much higher than that of \triple{elephant, CapableOf, climb tree} (score = $-0.839$, ranked 638,048 globally).

\paragraph{Text search}
Finally, we can use materialized resources for text search. For example, we can search in GPT2-XL-\conceptnet{} for all assertions that include the term \textit{``airplane''}, finding expected matches like \triple{airplane, AtLocation, airport} and \triple{flight attendant, CapableOf, travel on airplane}, as well as surprising ones like \triple{scrap paper, UsedFor, making paper airplane} and \triple{traveling, HasSubevent, sleeping on airplane}.
% \sr{can you build an example along these lines?}

\begin{table}[t]
    \centering
    \scriptsize
    \begin{tabular}{lp{0.62\columnwidth}}
    \toprule
    \textbf{Predicate} & \textbf{Most common objects} \\
    \midrule
    AtLocation & desk (3210), cabinet (2481), sock drawer (1771) \\
    \midrule
    CapableOf & branch out (963), branch off (747), taste good (556) \\
    \midrule
    Causes & death (2504), tears (1290), happiness (1254) \\
    \midrule
    Desires & eat (949), have fun (816), sex (742) \\
    \midrule
    HasA & more than one meaning (1387), seeds (1316), peel (1170) \\
    \midrule
    HasPrerequisite & metal (1965), plastic (1594), water (1423) \\
    \midrule
    HasProperty & good (2615), useful (2585), good for (1746) \\
    \midrule
    HasSubevent & breathe (1006), swallow (721), take off shoes (658) \\
    \midrule
    MadeOf & plastic (1427), aluminum (1297), wood (905) \\
    \midrule
    MotivatedByGoal & have fun (994), enjoyment (493), succeed (444) \\
    \midrule
    PartOf & new testament (914), human experience (683), alabama (667) \\
    \midrule
    ReceivesAction & found in house (1110), eaten (800), found in hospital (779) \\
    \midrule
    UsedFor & cooking (627), decoration (454), transport (448) \\
    \bottomrule
    \end{tabular}
    \caption{Most common objects generated by GPT2-XL-\conceptnet{}. Numbers in parentheses indicate frequency of the corresponding objects.}
    \label{tab:common-objects}
\end{table}

\section{Conclusion}

We introduced four CSKBs computed using two COMET models (BART and GPT2-XL) trained on two existing CSK resources (\conceptnet{} and \ascentpp{}). Our findings are:
% Main takeaways
% 1. Comet methodology produces better results on small, highest-quality data (ConceptNet) instead of larger good resource (Ascent++)
\begin{enumerate}
    \item The \comet{} methodology produces better results on modest manually curated resources (\conceptnet{}) than on larger web-extracted resources (\ascentpp{}).
    \item \comet{}'s recall can significantly outperform that of modest manually curated ones (\conceptnet{}), and reach that of large web-extracted ones (\ascentpp{}).
    \item In terms of precision, a significant gap remains to manual curation, both in typicality and saliency. To web extraction, a moderate gap remains in terms of statement typicality.
\end{enumerate}
%In the intrinsic evaluation, we showed that the \comet{} methodology produces better results on small, highest-quality data (\conceptnet{}) instead of a larger good resource extracted automatically from texts (\ascentpp{}).
% 2. Ascent++ has an edge over Comet, though not a big one (only for typicality, saliency on par). Though apart from numbers, it always has an edge in explainability.
%Although both leverage very large corpora as input,
%\ascentpp{} still has an edge over \comet{} as it demonstrated better typicality of CSK assertions, and more importantly, it always gives better explainability compared to pretrained LMs.
% 3. Some common COMET errors are ...., which might be overcome by more efforts on ... (post-filtering?)
We also identified common problems of the \comet{} generations, such as co-occurrence misreadings, subject copying, and redundancies, which may be subject of further research regarding post-filtering and clustering. 
% The compiled CSKBs are available for browsing and download at {\small \url{https://ascentpp.mpi-inf.mpg.de/comet}}.
% \sr{add screenshot of interface}

% \section{TODO}

% Table dataset size\\
% Table dataset examples\\
% Table precision/recall values\\

% [Li et al., 2016] Xiang Li, Aynaz Taheri, Lifu Tu, and Kevin
% Gimpel. Commonsense knowledge base completion. In
% Proceedings of ACL 2016, pages 1445–1455, 2016.

\bibliography{references}
\bibliographystyle{acl_natbib}

\end{document}